\RequirePackage{fix-cm}
\documentclass[smallextended]{svjour3}       
\smartqed  
\usepackage{graphicx}
\usepackage{booktabs}
\usepackage{float}
\usepackage{subfig}
\usepackage{cite}
\usepackage{url}
\usepackage{amssymb}
\usepackage{bm}
\begin{document}

\title{Auto-weighted Mutli-view Sparse Reconstructive Embedding
}


\author{Huibing Wang         \and
        Haohao Li         \and
        Xianping Fu \footnote{Corresponding Author} 
}


\institute{H. Wang \at
              College of Information and Science Technology, Dalian Maritime University, Dalian, China, 116021\\
              \email{huibing.wang@dlmu.edu.cn}           
           \and
           H. Li \at
              School of Mathematical Sciences, Dalian University of Technology, Dalian, China, 116024\\
              \email{haohaoli@mail.dlut.edu.cn} 
          \and
           X. Fu \at
              College of Information and Science Technology, Dalian Maritime University, Dalian, China, 116021\\
              \email{fxp@dlmu.edu.cn} 
}

\date{Received: date / Accepted: date}

\maketitle

\begin{abstract}
With the development of multimedia era, multi-view data is generated in various fields. Contrast with those single-view data, multi-view data brings more useful information and should be carefully excavated. Therefore, it is essential to fully exploit the complementary information embedded in multiple views to enhance the performances of many tasks. Especially for those high-dimensional data, how to develop a multi-view dimension reduction algorithm to obtain the low-dimensional representations is of vital importance but chanllenging. In this paper, we propose a novel multi-view dimensional reduction algorithm named Auto-weighted Mutli-view Sparse Reconstructive Embedding (AMSRE) to deal with this problem. AMSRE fully exploits the sparse reconstructive correlations between features from multiple views. Furthermore, it is equipped with an auto-weighted technique to treat multiple views discriminatively according to their contributions. Various experiments have verified the excellent performances of the proposed AMSRE.

\end{abstract}

\section{Introduction}
\label{intro}
Nowdays, we have witnessed the rapid development of information technology \cite{hu2017deep,feng2018learning,wang2018beyond}. It is common that one sample can be described from multiple perspectives, which leads to the large-scale multi-view data produced in various fields \cite{shen2015supervised}. Multi-view data not only contains more compatible and complementary information, but also improves the performances of those decision making systems \cite{wu2018and}. For example, one image can be represented by features extracted from multiple descriptors, such as, Local Binary Patterns (LBP) \cite{ahonen2004face}, Scale-Invariant Feature Transform (SIFT) \cite{ng2003sift} and Locality-constrained Linear Coding (LLC) \cite{wang2010locality}, etc \cite{feng2017spectral}.  All these features should be carefully exploited by multi-view learning algorithms. Therefore, researchers all over the world pay more attentions in the field of multi-view learning and develop various algorithms to meet the requirement of some applications \cite{wu2019cycle}.  

During the past decade, there are many multi-view learning algorithms \cite{wang2017unsupervised,Wang2017Effective} proposed using various techniques. Most multi-view learning algorithms focus on the task of clustering. Kumar et al. \cite{kumar2011co} proposed a co-regularized framework which can minimize the distinctions between multiple views. And it has achieved good performance to deal with multi-view clustering. Xia et al. \cite{xia2010multiview} has developed a auto learning trick to learn the factors corresponding to all views and combined graphs from multiple views. The proposed MSE has also attracted attentions from researchers in this field. Wang et al. \cite{wang2018multiview} finished the task of subspace clustering via structured low-rank matrix factorization and also achieved good performance. Moreover, there are some algorithms proposed to construct low-dimensional subspace \cite{wang2016multi} for multi-view data. Kan et al. \cite{kan2016multi} extended Linear Discriminant Analysis (LDA) \cite{mika1999fisher} into multi-view mode and proposed a method called Multi-view Discriminant Analysis (MvDA). Luo et al. \cite{luo2015tensor} extended canonical correlation analysis to the tensor mode, which can deal with multi-view data in tensor form and finish the task of dimension reduction. All these methods are proposed from different perspectives to deal with multi-view data \cite{wang2015lbmch}.

Meanwhile, high-dimensional data \cite{wu2018deep} has caused many problems to many applications, such as metric learning \cite{shen2011scalable,wang2016semantic}, face alignment \cite{liu2018face}, et al \cite{deng2018learning,wu20183d}. Therefore, how to obtain low-dimensional representations for high-dimensional features is also a hot topic in the last decades. Principle Component Analysis (PCA) \cite{Agarwal2009Face} and LDA \cite{mika1999fisher} are two most traditional ones in this fields. PCA is an unsupervised method which maximizes the global variance  of data to obtain the low-dimensional subspace. Even though it is simple and convenient, it lacks discriminative ability since it can not fully utilized enough information. LDA is a supervised method and fully utilizes label information. It has been utilized in many classification tasks because of it's ability. Locality Preserving Projections (LPP) \cite{he2004locality} is a local DR method which considers the relationships between each two neighbours and maintained them in the low-dimensional subspace. Neighborhood Preserving Embedding (NPE) \cite{he2005neighborhood} is another local DR method which maintained the linear reconstructive relationships between samples. Sparsity Preserving Projection \cite{qiao2010sparsity} is a DR method which exploits the sparse relationships between samples. All these methods are proposed to construct low-dimensional subspace for high-dimensional data, which has attracted wide attentions \cite{wu2018whatand} from authors all over the world. 

In this paper, we focused on constructing the low-dimensional representations for multi-view data and proposed a novel method named Auto-weighted Mutli-view Sparse Reconstructive Embedding. Because multiple views have different impacts on the algorithm, AMSRE can automatically assign differnet factors to multiple views according their contributions. Furthermore, AMSRE fully exploited the sparse reconstructive relationships between features within their perspective views. Then, AMSRE maintained the relationships and forced all views to help each other to improve its discriminative ability. The overall framework of AMSRE has been shown as Fig.\ref{fig1}. And we summarized the contributions of AMSRE as follows:

\begin{figure}[htbp]
\label{fig1}
\centering
\includegraphics[width=\textwidth]{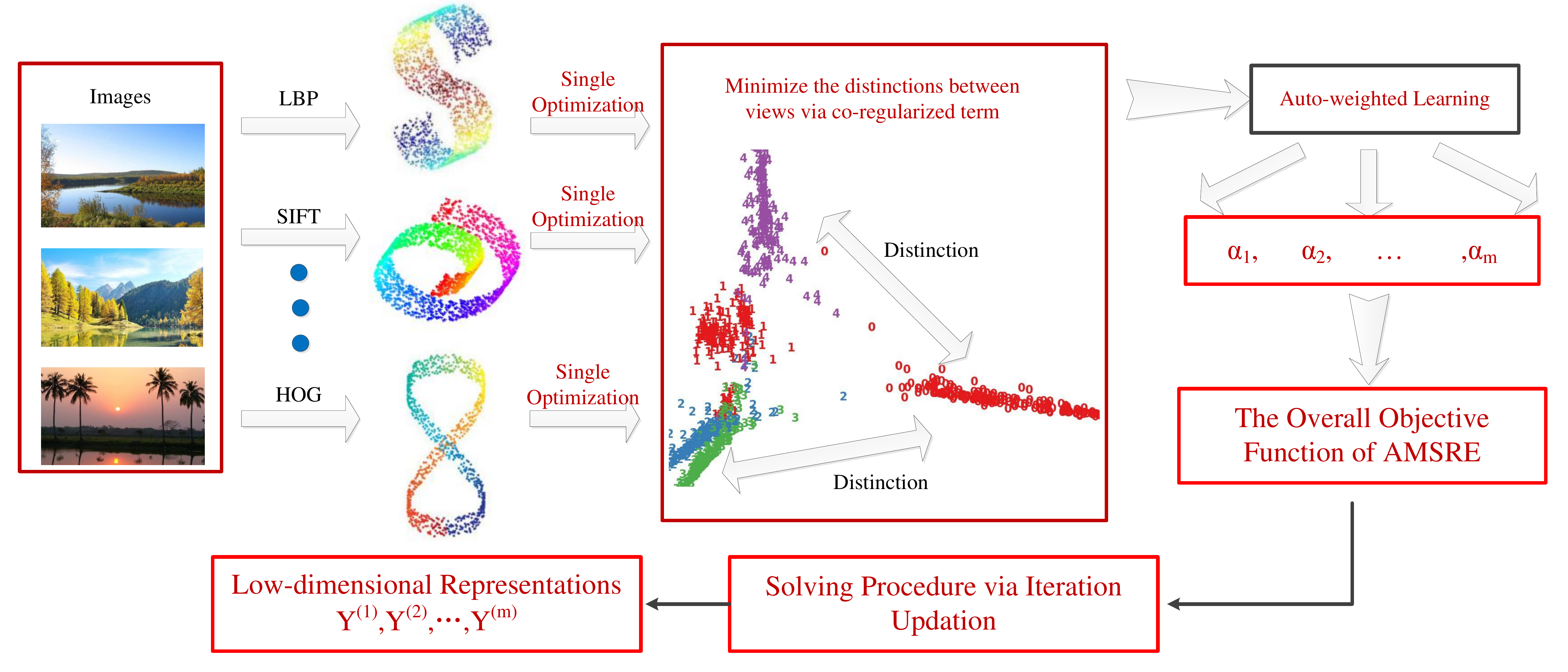}
\caption{ The working procedure of AMSRE.}
\end{figure}

\subsection{Constructing Procedure}

\begin{itemize}
    \item AMSRE is successfully equipped with a auto-weighted method to assign multiple views with different factors. This procedure can help AMSRE better understands the contributions of different views.
    
    \item AMSRE can better maintain the spares reconstructive relationships between features within their perspective views, which can improve the discriminative ability of the low-dimensional representations.
    
    \item We carefully construct an alternating optimization method to obtain the solution of AMSRE, which can be refered by some related studies.
\end{itemize}

The following paper is organized as follows: in section 2, we introduced the basic knowledge of multi-view learning and summarized some related works in this field. In section 3, we illustrated the construction process of AMSRE and described the solving procedure in detail. Section 4 shown various experiments to verify the performance of our proposed AMSRE. And we made a conclusion of this paper in section 5.

\section{Related Works}

In this section, we introduced some basic knowledge of multi-view learning \cite{Wang2016Iterative}. Furthermore, we have shown 2 typical multi-view learning methods.

Assume we are given a multi-view dataset $\bm{X} = \left\{ \bm{X}^v \in \Re^{D_v \times N}, v=1,\cdots,m \right\}$ which contains $N$ samples from $m$ views. $\bm{X}^v$ consists of $N$ features in the $v$th view. All features in the $v$th view locate in a $D_v$-dimensional space. Multi-view learning is an essential research field to fully utilize information from multiple views to obtain a better decision. Therefore, the goal of our proposed AMSRE is to construct a common subspace for features from all views and obtain the low-dimensional representations $\bm{Y} = \left\{ \bm{Y}^v \in \Re^{d \times N} \right\}$ for the original multi-view data, where $d<D_v, v=1,\cdots,m$.

\subsection{Multiview Spectral Embedding}

MSE is a good performance for multi-view dimension reduction. It can encode different features from multiple views to achieve a physically
meaningful embedding. Xia el al. \cite{xia2010multiview} extends Laplacian Eigenmaps (LE) \cite{belkin2002laplacian} into multi-view mode and develops an architecture to learn weights for different views according to their contributions. Furthermore, MSE integrates laplacian graphs from multiple views via global coordinate alignment. And the proposed objective function of MSE can be summarized as follows:

\begin{equation}
\label{eq1}
\begin{array}{l}
 \mathop {\arg \min }\limits_{\alpha ,Y} \sum\limits_{v = 1}^m {\alpha _v
tr\left( {YL^{\left( v \right)}Y^T} \right)} \\
 s.t.\;YY^T = I;\;\sum\limits_{v = 1}^m {\alpha _v } = 1,\;\;\alpha _v \ge 0
\\
 \end{array}
\end{equation}

where $L^{\left( v \right)}$ is the laplacian graph for features in the $v$th view. It reflects the neighborhood relationship between
features in the $v$th view. $\alpha = \left[ {\alpha _1 ,\alpha _2 , \cdots ,\alpha _m } \right]$ is a set of coefficients which can reflect the importance of different views. And $Y$ is the low-dimensional representation for the original multi-view data. And MSE develops an iterative optimization procedure to update $\alpha$ and $Y$ alternately.

\subsection{Co-regularized Multi-view Spectral Clustering}

Co-regularized Multi-view Spectral Clustering \cite{kumar2011co} is a novel multi-view method to deal with the task of clustering \cite{wang2015robust}. It first utilized a co-regularized term to minimize the distinctions between multiple views and calculated the low-dimensional representations for all samples in each view. Then, traditional spectral clustering strategy can be carried on to assign all samples into different clusters. And an iterative optimization procedure is adopted to solve the solution of this method. The objective function is shown as follows:

\begin{equation}
\label{eq2}
\begin{array}{l}
\mathop {\max }\limits_{{U^{(1)}},{U^{(2)}}, \cdots ,{U^{(m)}}} \sum\limits_{v = 1}^m {tr\left( {{U^{{{(v)}^T}}}{L^{(v)}}{U^{\left( v \right)}}} \right)}  + \lambda \sum\limits_{1 \le v \ne w \le m}^{} {tr\left( {{U^{\left( v \right)}}{U^{{{(v)}^T}}}{U^{\left( w \right)}}{U^{{{(w)}^T}}}} \right)} \\
s.t.\;\;\;\;{U^{\left( v \right)}}{U^{{{(v)}^T}}}{\rm{ = }}I,\;\;\;\forall 1 \le v \le m
\end{array}
\end{equation}

where $U^{{{(v)}}}$ is the low-dimensional representation for features in the $v$th view. $L^{(v)}$ is the laplacian graph for the $v$the view. $\lambda$ is a regularized parameter to balance the weights of each two views. The second term in Eq.\ref{eq2} can minimize the distinctions between each two views to help them to learn from each other to obtain the low-dimensional representations. 

\section{The Proposed Method}

\subsection{The Construction Process of AMSRE}

In this section, we introduced the proposed Auto-weighted Mutli-view Sparse Reconstructive Embedding (AMSRE) in detail. AMSRE aims to integrate compatible and complementary information from multiple views and utilized the co-regularized term to minimize the distinctions between all views. Furthermore, AMSRE is equipped with a auto-weighted strategy to assign factors to each views according to their contributions. Therefore, the obtained low-dimensional representation can better maintain information from multi-view data.  First, we aim to maintain the sparse reconstructive correlations in the $v$th view as follows:

\begin{equation}
\label{eq3}
\arg\mathop {\min }\limits_{Y^{(v)}} \quad \sum_{i=1}^n ||y_i^{(v)} - Y_i^{(v)}  s_i^{(v)}||^2
\end{equation}

where $Y_i^{(v)} $ is the set of features in the $v$th view, which has not contain $y_i^{(v)}$.      $s_{i}^{(v)}$ is the sparse reconstructive correlation vector which can be calculated by sparse representation \cite{qiao2010sparsity}. Eq.\ref{eq3} aims to construct the low-dimensional representation  $Y_i^{(v)} $ for  $X_i^{(v)} $ which can contain sparse reconstructive correlations in the original multi-view data. According to mathematical transformation, Eq.\ref{eq3} can be expressed as follows:
 
\begin{equation}
\label{eq4} 
\begin{array}{l}
 \mathop {\arg \min }\limits_{{Y^{(v)}}} tr\left( {{{\left( {{Y^{\left( v \right)}}} \right)}^T}{M^{\left( v \right)}}{Y^{\left( v \right)}}} \right)\\
s.t. \left(Y^{(v)}\right)^T Y^{(v)} = I, v=1,2,\cdots,m.
\end{array}
\end{equation}

where $M^{(v)} = (I-S^{(v)})(I-S^{(v)})^T$ and $S^{(v)} = \left[s_1^{(v)},s_2^{(v)},\cdots,s_n^{(v)}\right]$. And $Y^{(v)}$ is the low-dimensional representation for features in the $v$th view. However, Eq.\ref{eq4} is the single view method which can only calculate for one single. In order to extend Eq.\ref{eq4}, we first minimize the sum of Eq.\ref{eq4} for all views as follows:

\begin{equation}
\label{eq5} 
\begin{array}{l}
\mathop {\arg \min }\limits_{Y^{(1)},Y^{(2)},\cdots,Y^{(m)}} \sum\limits_{v=1}^m tr\left( {{{\left( {{Y^{\left( v \right)}}} \right)}^T}{M^{\left( v \right)}}{Y^{\left( v \right)}}} \right)\\
s.t. \left(Y^{(v)}\right)^T Y^{(v)} = I, v=1,2,\cdots,m.
\end{array}
\end{equation}

Even though Eq.\ref{eq5} take all views into considerations, it cannot help all views to learn from each other. Therefore, we introduced a co-regularized term to minimize the distinctions between all views. We propose the following cost function as a measure of disagreement between each two views:

\begin{equation}
\label{eq6} 
D\left( {{Y^{\left( v \right)}},{Y^{\left( w \right)}}} \right) = \left\| {\frac{{{K_{{Y^{\left( v \right)}}}}}}{{\left\| {{K_{{Y^{\left( v \right)}}}}} \right\|_F^2}} - \frac{{{K_{{Y^{\left( w \right)}}}}}}{{\left\| {{K_{{Y^{\left( w \right)}}}}} \right\|_F^2}}} \right\|_F^2
\end{equation}

where $K_{{Y^{\left( v \right)}}}$ is the similarity matrix for $Y^{\left( v \right)}$, and $\left\|  \bullet  \right\|_F$ denotes the Frobenius norm of the matrix. Eq.\ref{eq6} can be utilized measure the disagreement between each two views. And minimizing Eq.\ref{eq6} can keep all views to be consensus. Because $K_{{Y^{\left( v \right)}}} =Y^{\left( v \right)}\left(Y^{\left( v \right)}\right)^T$, Eq.\ref{eq6} can be further transformed as follows:

\begin{equation}
\label{eq7} 
D\left( {{Y^{\left( v \right)}},{Y^{\left( w \right)}}} \right) =  - tr\left( {{Y^{\left( v \right)}}{{\left( {{Y^{\left( v \right)}}} \right)}^T}{Y^{\left( w \right)}}{{\left( {{Y^{\left( w \right)}}} \right)}^T}} \right)
\end{equation}

The transform from Eq.\ref{eq6} to Eq.\ref{eq7} neglects constant additive and scaling terms. Therefore, combines with Eq.\ref{eq7}, The objective function of AMSRE can be organized as

\begin{equation}
\label{eq8} 
\begin{array}{l}
\mathop {\arg \min }\limits_{Y^{(1)},Y^{(2)},\cdots,Y^{(m)}} \sum\limits_{v=1}^m tr\left( {{{\left( {{Y^{\left( v \right)}}} \right)}^T}{M^{\left( v \right)}}{Y^{\left( v \right)}}} \right) \\
~~~~~~~~~~~~~~~  +  \lambda \sum\limits_{1 \le v \ne w \le m}^{} {tr\left( {{Y^{\left( v \right)}}{{\left( {{Y^{\left( v \right)}}} \right)}^T}{Y^{\left( w \right)}}{{\left( {{Y^{\left( w \right)}}} \right)}^T}} \right)}\\
s.t. \left(Y^{(v)}\right)^T Y^{(v)} = I, v=1,2,\cdots,m.
\end{array}
\end{equation}

It is clear that we can obtain the low-dimensional representations through Eq.\ref{eq8}. However, because multiple views have different influences on the construction of low-dimensional representations. Therefore, we should further exploit information in different views and assign different weights to different views. Therefore, we equip an auto-weighted trick with Eq.\ref{eq8} and reformulate the objective function of AMSRE as follows:s

\begin{equation}
\label{eq9} 
\begin{array}{l}
\mathop {\arg \min }\limits_{Y^{(1)},Y^{(2)},\cdots,Y^{(m)},\alpha}  \sum\limits_{v=1}^m \alpha_v^r tr\left( {{{\left( {{Y^{\left( v \right)}}} \right)}^T}{M^{\left( v \right)}}{Y^{\left( v \right)}}} \right)  \\
~~~~~~~~~~~~~~~  + \lambda \sum\limits_{1 \le v \ne w \le m}^{} {tr\left( {{Y^{\left( v \right)}}{{\left( {{Y^{\left( v \right)}}} \right)}^T}{Y^{\left( w \right)}}{{\left( {{Y^{\left( w \right)}}} \right)}^T}} \right)}\\
s.t. \left(Y^{(v)}\right)^T Y^{(v)} = I, v=1,2,\cdots,m.\\
~~~~  \sum_{v=1}^m \alpha_v = 1
\end{array}
\end{equation}

where $\alpha_v$ is the weight to reflect the importance of the $v$th view. $\alpha = [\alpha_1,\alpha2,\cdots,\alpha_m]$ is the weight vector. And the low-representations $Y^{(v)}$ in Eq.\ref{eq9} can be calculated by eigen-decomposition. And we provide the solving process of AMSRE in the following section.

\subsection{Solving Procedure of AMSRE}

We have shown how we construct the objective funciton of AMSRE before. In this section, we provide the solving process of it. Because AMSRE should optimize $Y^{(v)},v=1,2,\cdots,m$ with $\alpha$ at the same time, we adopts an iterative optimization strategy to obtain the solution. For each iteration, if we want to update $Y^{(v)}$, we should maintain all the other variables to be unchanged, including $Y^{(i)},v=1,2,\cdots,v-1,v+1,\cdots,m$ and $\alpha$. Therefore, the objective function of AMSRE can be organized as follows:

\begin{equation}
\label{eq10} 
\begin{array}{l}
\mathop {\arg \min }\limits_{Y^{(v)}}  \alpha_v^r tr\left( {{{\left( {{Y^{\left( v \right)}}} \right)}^T}{M^{\left( v \right)}}{Y^{\left( v \right)}}} \right)  \\
~~~~~~~~~~~~~~~  + \lambda \sum\limits_{w \ne v }^{} {tr\left( {{Y^{\left( v \right)}}{{\left( {{Y^{\left( v \right)}}} \right)}^T}{Y^{\left( w \right)}}{{\left( {{Y^{\left( w \right)}}} \right)}^T}} \right)}\\
s.t. \left(Y^{(v)}\right)^T Y^{(v)} = I

\end{array}
\end{equation}

Due to the additive operation of trace, Eq.\ref{eq10} can be further transformed as

\begin{equation}
\label{eq11} 
\begin{array}{l}
\mathop {\arg \min }\limits_{Y^{(v)}}   tr \left( {{\left( {{Y^{\left( v \right)}}} \right)}^T}\left(\alpha_v^r {M^{\left( v \right)}}+\lambda Y^{\left( w \right)}{\left( {{Y^{\left( w \right)}}} \right)}^T \right){Y^{\left( v \right)}} \right) \\
s.t. \left(Y^{(v)}\right)^T Y^{(v)} = I

\end{array}
\end{equation}

Therefore, we can get the low-dimensional representation  $Y^{(v)}$ by calculating the eigenvector of $\alpha_v^r {M^{\left( v \right)}}+\lambda Y^{\left( w \right)}{\left( {{Y^{\left( w \right)}}} \right)}^T $ with the constraint $\left(Y^{(v)}\right)^T Y^{(v)} = I$. We can update all the low-dimensional representations $Y^{(v)},v=1,2,\cdots,m$ by keep the other variable unchanged and just update one view.

Meanwhile, in order to obtain $\alpha$, we adopt Lagrange multiplier to update it. After we update all the  $Y^{(v)},v=1,2,\cdots,m$, we keep them unchanged and update $\alpha$. . By using a Lagrange multiplier $\eta$ to take the constraint $\sum_{v=1}^m \alpha_v = 1$ into consideration, we get the Lagrange function as

\begin{equation}
\label{eq12} 
\begin{array}{l}
L\left(\alpha,\eta \right) =\mathop {\arg \min }\limits_{Y^{(v)}}  \alpha_v^r tr\left( {{{\left( {{Y^{\left( v \right)}}} \right)}^T}{M^{\left( v \right)}}{Y^{\left( v \right)}}} \right)  + \eta \left(\sum\limits_{v=1}^m \alpha_v = 1\right)
\end{array}
\end{equation}

By setting the derivative of $L\left(\alpha,\eta \right)$ with respect to $\alpha_v$ and $\eta$ to zero, we have

\begin{equation}
\label{eq13}
    \left\{ { \frac{\partial L\left(\alpha,\eta \right)}{\partial \alpha_v} = r \alpha_v^{\left(r-1\right)} tr\left( {{{\left( {{Y^{\left( v \right)}}} \right)}^T}{M^{\left( v \right)}}{Y^{\left( v \right)}}} \right) - \eta = 0, v =1,2,\cdots,m. \atop \frac{\partial L\left(\alpha,\eta \right)}{\partial \eta} =\sum\limits_{v=1}^m \alpha_v - 1 =0  ~~~~~~~~~~~~~~~~~~~~~~~~~~~~~~~~~~~~~~~~~~~~~~~~~~~~} \right.
\end{equation}

Therefore, $\alpha_v$ can be update by the following rules. 

\begin{equation}
\label{eq14}
\alpha_v = \frac{1/tr\left( {{{\left( {{Y^{\left( v \right)}}} \right)}^T}{M^{\left( v \right)}}{Y^{\left( v \right)}}}   \right)^{1/\left(r-1\right)}}{\sum\limits_{v=1}^{m}\left( 1/ {{{\left( {{Y^{\left( v \right)}}} \right)}^T}{M^{\left( v \right)}}{Y^{\left( v \right)}}}   \right)^{1/\left(r-1\right)}}  
\end{equation}

It can be calculated by Eq.\ref{eq14} to update $\alpha$. And we can obtain the optimal  $Y^{(v)},v=1,2,\cdots,m$ and $\alpha$ by updating one of them and keeping the other $m$ variables unchanged. And we conclude the solving procedure in Table.\ref{tab1}.

\begin{table}[ht]
\caption{The optimization procedure of AMSRE}
\label{tab1}
\begin{tabular}
{p{300pt}}
\toprule[2pt]
\textbf{Input:} \\
        A set of multi-view features with $N$ training samples having $m$ views $X^{\left( v \right)} = \left[ {x_1^{\left( v \right)} ,x_2^{\left( v \right)} , \cdots ,x_N^{\left( v \right)} } \right] \in R^{D_{v} }$. \\
\textbf{Initialization:} \\
 Initialize  $Y^{(v)},v=1,2,\cdots,m$ using single view optimization as Eq.\ref{eq4} \\
\textbf{The optimization procedure of AMSRE:} \\
1. Do  \\
2. ~~~~Using sparse representation to construct the sparse reconstructive weights\\
~~~~~~~ matrix $S^{\left( v \right)},v = 1,2, \cdots ,m$ for all views . \\
3. ~~~~Calculate $ M^{(v)} = (I-S^{(v)})(I-S^{(v)})^T ,v = 1,2, \cdots ,m$ for all views. \\
5. ~~~~For $v = 1:m$ \\
6.~~~~~~~~ Update $Y^{\left( v \right)}$ for the $v$th view according to Eq.(\ref{eq11}) \\
7. ~~~~End \\
8. ~~~~Update $\alpha$ according to Eq.(\ref{eq14}) \\
9. Until $Y^{(v)},v=1,2,\cdots,m$ converges \\
 \textbf{Output:} \\
 The low-dimensional representation $Y^{(v)},v=1,2,\cdots,m$ for all views \\
\bottomrule[2pt]
\end{tabular}
\end{table}

\section{Experiment}

In this section, we conduct several experiments on the benchmark multi-view datasets (including 3Sources, Cora, WebKB, Yale and ORL)   to verify the performance of our proposed AMSRE. First, we introduced the utilized datasets in this section and listed some comparing methods. Then, we carry on experiments on these datasets and provide the results on them. 

\subsection{Datasets and Comparing Methods}

\begin{figure}[htbp]
\label{fig2}
\centering
\includegraphics[width=\textwidth]{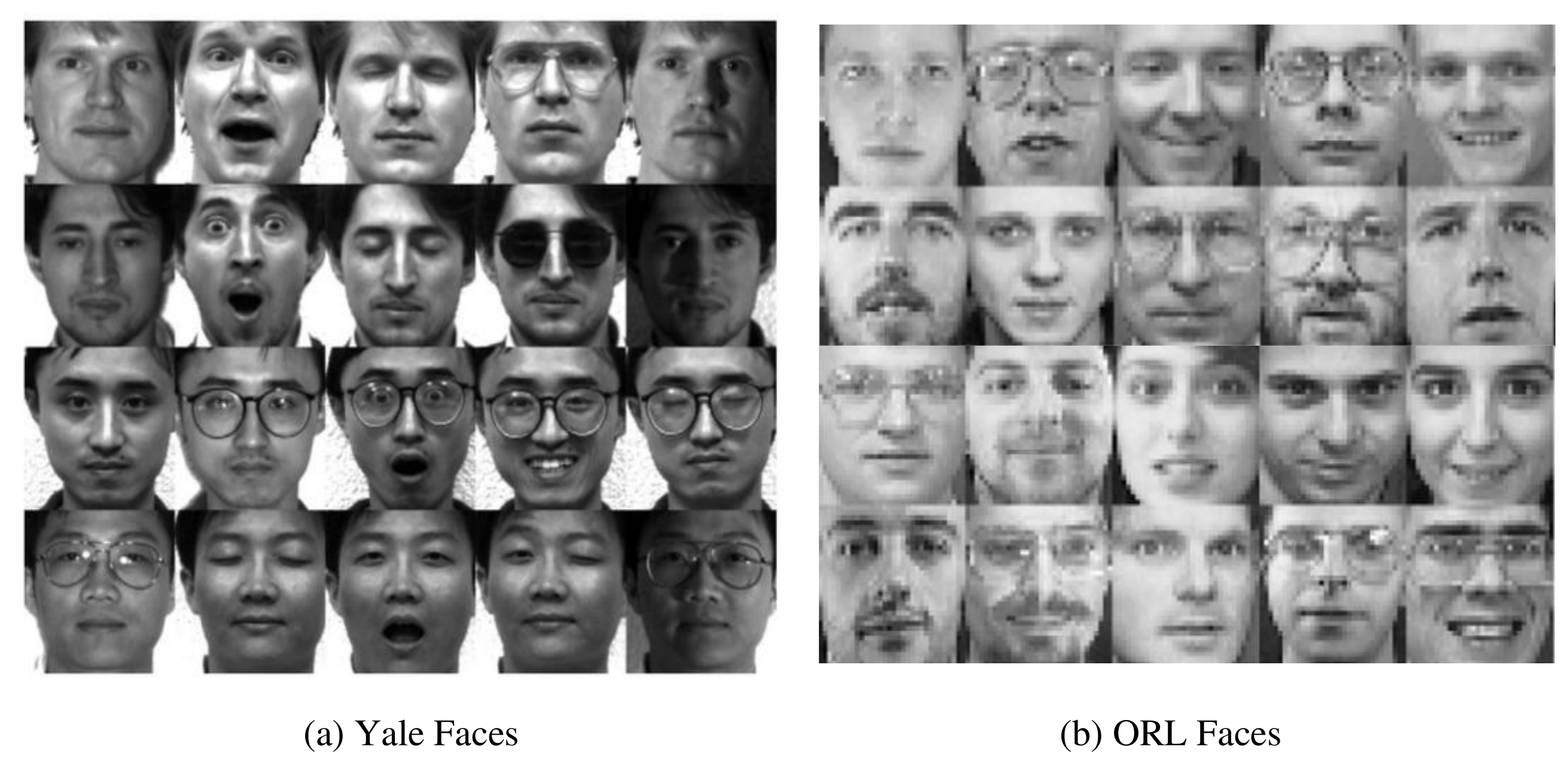}
\caption{ Some faces from Yale and ORL datasets}
\end{figure}

In our experiments, 5 datasets are utilized to illustrate the effectiveness of AMSRE, including document datasets (3sources \footnote{http://mlg.ucd.ie/datasets/3sources.html}, Cora \footnote{https://relational.fit.cvut.cz/dataset/CORA} and WebKB \footnote{http://www.webkb.org/}) and face datasets (Yale \footnote{http://cvc.cs.yale.edu/cvc/projects/yalefaces/yalefaces.html} and ORL \footnote{https://www.cl.cam.ac.uk/research/dtg/attarchive/facedatabase.html}). For those images datasets, we extract features using multiple descriptors as multi-view features for our experiments,which has been shown in the corresponding experiments. Some images from these datasets are shown as Fig.\ref{fig2}.

We adopt the following methods as comparing ones: 1. Co-reg \cite{kumar2011co}, 2 .Canonical correlation analysis (CCA) \cite{hardoon2004canonical}, 3. Sparsity preserving projections (SPP) \cite{qiao2010sparsity}, 4. Multiview spectral embedding (MSE) \cite{xia2010multiview}. We project multi-view data into low-dimensional subspace and then using 1NN \cite{denoeux1995k} to test all the performance of the comparing methods and AMSRE. We calculated all the experiment results on the low-dimensional representations from each single view. And the experiment results are the best ones from all views. All samples from each dataset are randomly separated as two parts (training set and testing set).

\subsection{Document Classification}

In this section, we conducted related experiments on 3 document datasets, including 3Sources, Cora and WebKB datasets. For 3 Sources, it is collected from three online new sources, BBC, Reuters and Guardian. Therefore, 3Sources consisits of features from 3 views and each source is viewed as one view of 3Sources. There are 169 samples which comes from 6 classes in total. The dimensions of features from these 3 views are 3068, 3631, 3560
respectively. In our experiment, we randomly select twenty percent samples as testing ones while the other samples are assigned as training ones. After dimension reduction by those methods,  we conduct this experiment for 20 times and calculated the mean and max classification accuracies as table.\ref{tab2}.

\begin{table}[htbp]
\centering
\caption{The classification accuracies on 3Sources dataset}
\begin{tabular}
{ccccccccc}
\toprule[2pt]
\textbf{3Sources}& &Co-reg \cite{kumar2011co} &CCA &SPP  &MSE &AMSRE \\
\hline
\raisebox{-1.50ex}[0cm][0cm]{Dim=10 }&Mean &  72.36{\%} &   70.98\%  & 69.93{\%} & 74.56{\%} &  \textbf{75.42{\%}}  \\

 &Max & 82.73{\%} & 82.64{\%} & 80.33{\%} &85.70{\%}  &  \textbf{86.73{\%}}  \\
\hline
\raisebox{-1.50ex}[0cm][0cm]{Dim=30 }&Mean &  75.49{\%} &   74.98\%  & 73.14{\%} & 76.49{\%} &  \textbf{78.47{\%}}  \\

 &Max & 86.19{\%} & 85.51{\%} & 83.87{\%} &88.03{\%}  &  \textbf{90.23{\%}}  \\
\hline
\raisebox{-1.50ex}[0cm][0cm]{Dim=50 }&Mean & 81.30{\%} & 80.02{\%}  & 78.93{\%} & 83.06{\%} & \textbf{85.73{\%}} \\
 &Max & 88.14{\%} &86.96{\%}  & 85.34{\%} & 88.34{\%} &  \textbf{91.44{\%}}  \\
\bottomrule[2pt]
\end{tabular}
\label{tab2}
\end{table}

We have projected multi-view data into subspaces with different dimensions (such as 10, 30, 50).  It can be found easily that AMSRE can achieve best performances in most situations. Only SPP is the single view DR method and it performs worst among all methods. Furthermore, MSE also performs well than the other methods. Therefore, AMSRE is a better multi-view DR methods and it can fully exploits sparse reconstructive correlations between features from multiple views.

Cora dataset is collected by 2708 scientific publications which come from 7 classes. Each document is represented by content and cites information. Therefore, Cora is a multi-view data which contains 2 views. In our experiment, we randomly select twenty percent samples as testing ones while the other samples are assigned as training ones. After dimension reduction by those methods,  we conduct this experiment for 20 times and calculated the mean and max classification accuracies as table.\ref{tab3}.
 
\begin{table}[htbp]
\centering
\caption{The classification accuracies on Cora dataset}
\begin{tabular}
{ccccccccc}
\toprule[2pt]
\textbf{Cora}& &Co-reg \cite{kumar2011co} &CCA &SPP  &MSE &AMSRE \\
\hline
\raisebox{-1.50ex}[0cm][0cm]{Dim=10 }&Mean & 44.37{\%} &42.11{\%}  &  39.51{\%} & 48.72{\%} & \textbf{51.80{\%}}  \\

 &Max & 56.37{\%}  & 53.49{\%}  & 46.17{\%}  & 57.11{\%} & \textbf{60.22{\%}} \\
\hline
\raisebox{-1.50ex}[0cm][0cm]{Dim=30 }&Mean & 48.33{\%} &46.58{\%}  &  41.20{\%} & 50.33{\%} & \textbf{53.86{\%} } \\

 &Max & 56.37{\%}  & 53.49{\%}  & 46.17{\%}  & 57.11{\%} & \textbf{60.22{\%}} \\
\hline
\raisebox{-1.50ex}[0cm][0cm]{Dim=50 }&Mean & 52.10{\%}  & 49.74{\%}   &  42.78{\%}  & 54.37{\%} & \textbf{56.49{\%}} \\
 &Max & 61.11{\%}  & 58.78{\%} & 45.77{\%}  & 63.54{\%} &  \textbf{66.03{\%}} \\
\bottomrule[2pt]
\end{tabular}
\label{tab3}
\end{table}

WebKB contains 4 subsets of documents over 6 labels. A web pages consists of the following information: the text on it, the anchor text on the
hyperlink pointing to it and the text in its title. Therefore, WebKB is a multi-view data which has 3 views. In our experiment, we randomly select twenty percent samples as testing ones while the other samples are assigned as training ones. After we project multi-view data into a 30-dimensional subspace, we calculated the mean and max classification accuracies as table.\ref{tab4}.

\begin{table}[htbp]
\label{tab4}
\centering
\small
\caption{The classification accuracies on WebKB dataset}
\begin{tabular}
{ccccccccc}
\toprule[2pt]
\raisebox{-1.50ex}[0cm][0cm]{WebKB}&
\multicolumn{2}{c}{WebKB-1} &
\multicolumn{2}{c}{WebKB-2} &
\multicolumn{2}{c}{WebKB-3} &
\multicolumn{2}{c}{WebKB-4}  \\
\cline{2-9}
 &
Mean& Max &Mean &Max &Mean &Max &Mean &Max \\
\cline{1-9}
Co-reg \cite{kumar2011co} &83.46{\%}&87.33{\%}&67.95{\%}&76.54{\%}&87.18{\%}&90.10{\%}&75.43{\%}&80.24{\%} \\

CCA&83.34{\%}&89.44{\%}&78.23{\%}&81.62{\%}&87.02{\%}&92.47{\%}&68.18{\%}&76.23{\%} \\

SPP&82.54{\%}&87.30{\%}&67.19{\%}&72.33{\%}&88.81{\%}&92.79{\%}&77.53{\%}&79.80{\%} \\

MSE&85.33{\%}&89.23{\%}&75.26{\%}&80.99{\%}&90.33{\%}&91.93{\%}&79.68{\%}&83.22{\%} \\

AMSRE&87.25{\%}&90.96{\%}&77.18{\%}&82.99{\%}&92.17{\%}&94.36{\%}&81.63{\%}&85.92{\%} \\
\bottomrule[2pt]
\end{tabular}
\end{table}

It can be found that our proposed AMSRE can achieve best performances compared with the other methods. Meanwhile, multi-view algorithms are better than single-view ones to deal with multi-view dataset. Even though some methods can also achieve good performances in some situations, our proposed AMSRE is the best one. It can exploit sparse reconstructive correlations maintained in multi-view data and assign different weights to multiple views according to their contributions, which are the reasons why AMSRE is the best one.

\subsection{Face Recognition}

\begin{figure}[htbp]
\label{fig3}
\centerline{
\subfloat{\includegraphics[width=2.5in]{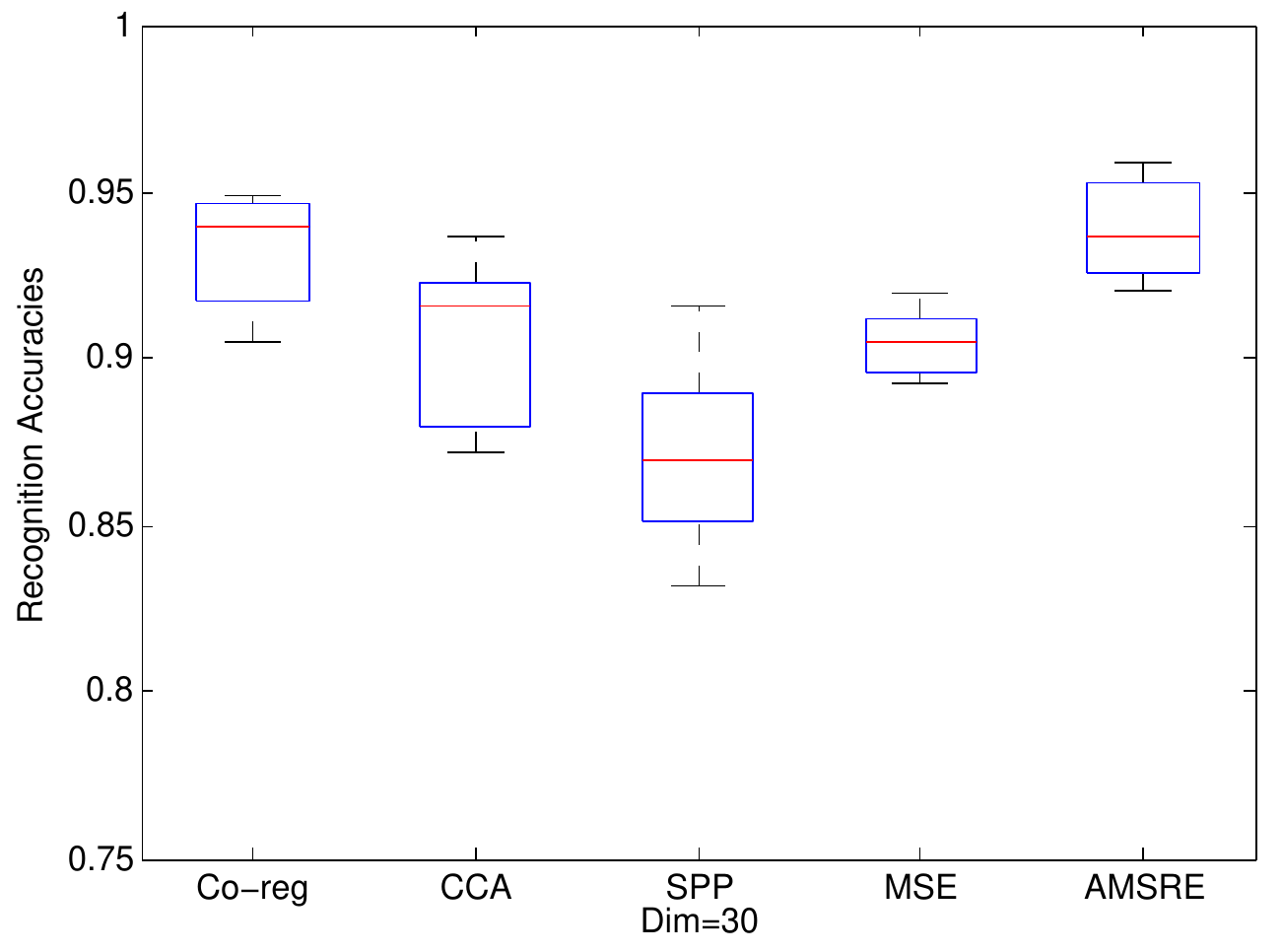}}
\subfloat{\includegraphics[width=2.5in]{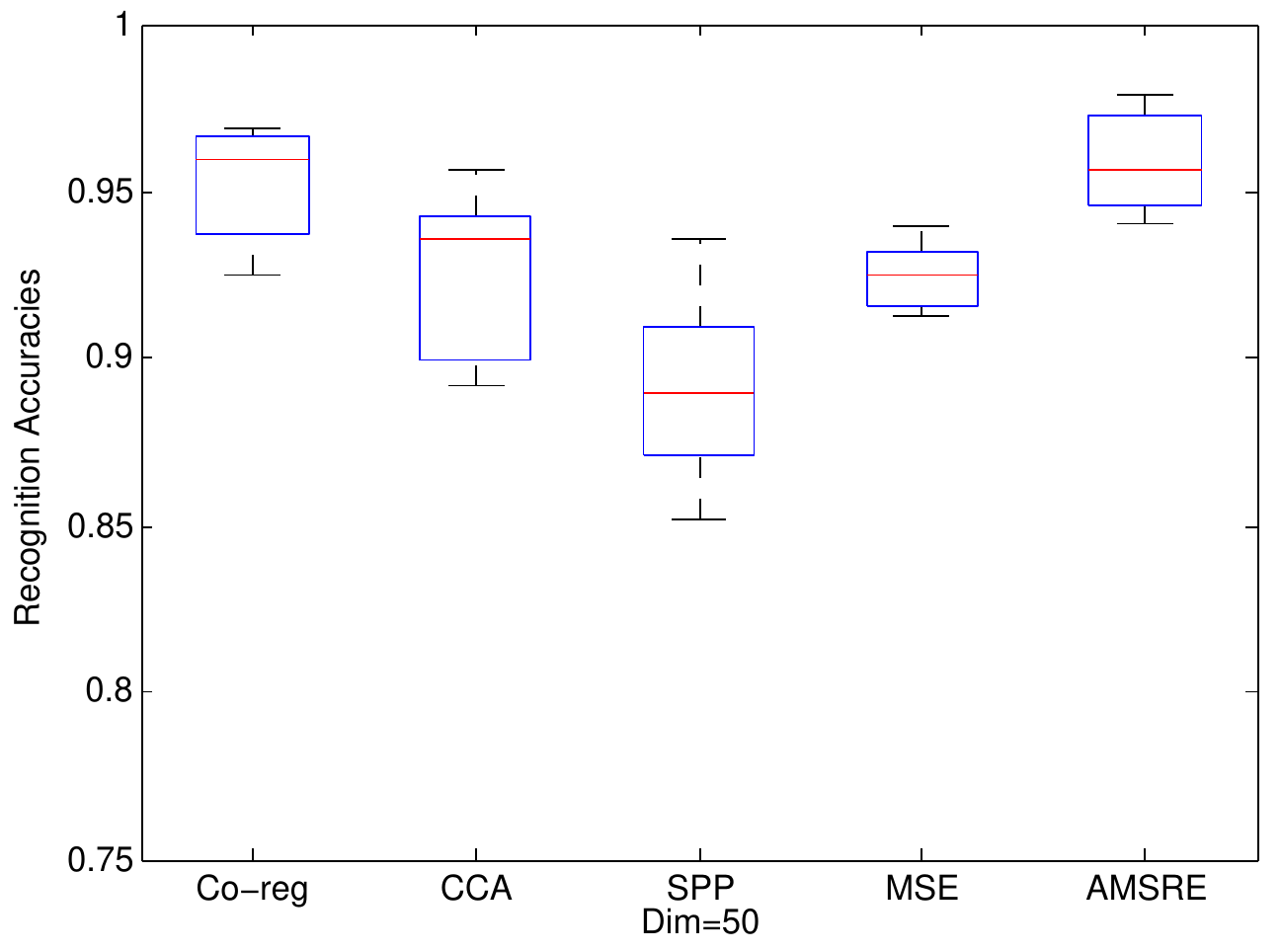}}
}
\caption{Recognition accuracies on Yale dataset in different dimensional subspaces}
\end{figure}

In this section, we construct some experiments on face recognition. We utilized 2 face datasets as experiments datasets and applies all DR methods on them. For Yale dataset, there are 165 faces corresponding to 11 people. We extract features by GSI \cite{fant1994grey}, LBP \cite{ahonen2004face} and EDH \cite{gao2008image} as three views. The dimensions of features from these 3 views are 1024, 256, 72 respectively. Similar with the experiments before, twenty percent samples are assigned as testing ones while the other faces are assign as training ones. 1NN classifier is adopted
to calculate the recognition results after the dimension reduction. And we show the experiments results in Fig.\ref{fig3}.

\begin{figure}[htbp]
\label{fig4}
\centerline{
\subfloat{\includegraphics[width=2.5in]{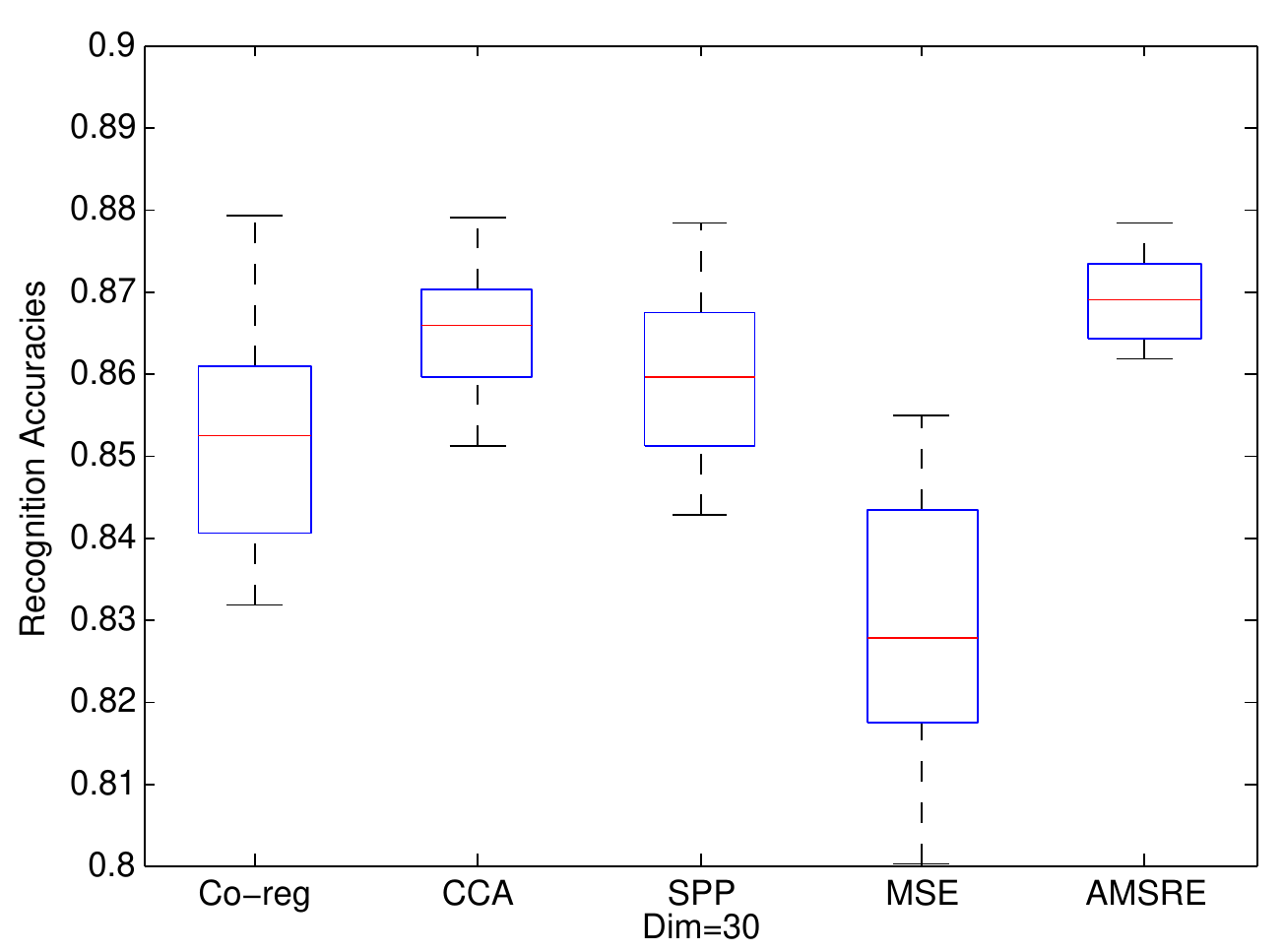}}
\subfloat{\includegraphics[width=2.5in]{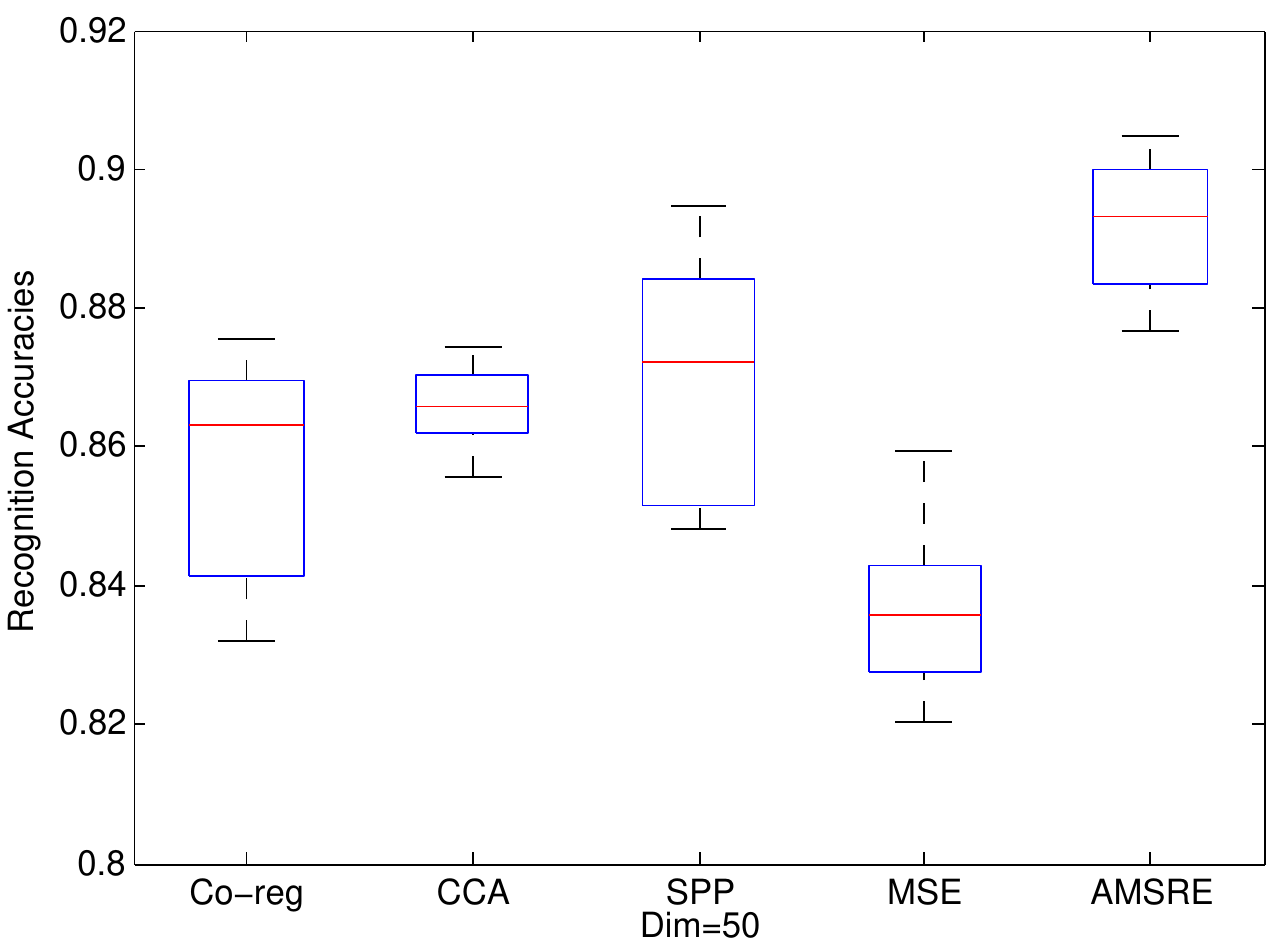}}
}

\caption{Recognition accuracies on ORL dataset in different dimensional subspaces}
\end{figure}

For ORL dataset, there are 400 faces corresponding to 40 people in total. We also extract features by GSI \cite{fant1994grey}, LBP \cite{ahonen2004face} and EDH \cite{gao2008image} as three views. The dimensions of features from these 3 views are 1024, 256, 72 respectively. twenty percent samples are assigned as testing ones while the other faces are assign as training ones. 1NN classifier is adopted
to calculate the recognition results after the dimension reduction. And we show the experiments results in Fig.\ref{fig4}.

We can also find that our proposed AMSRE can achieve best performances in Yale and ORL face datasets. Furthermore, the performances of multi-view DR methods are better. Because AMSRE fully exploits sparse reconstructive correlations between samples, it can better maintain information from multi-view data.

\section{Conclusion}

In this section, we proposed a novel multi-view DR method named AMSRE. It can fully exploit sparse reconstructive correlations between features from multiple views. Furthermore, it develops a technique to integrate multi-view information together and adopts a auto-weighted learning method which can assign multiple views with different weights according to their contributions. We have conducted several experiments to verify the performance of our proposed AMSRE. And it can achieve excellent performances in most situations. 

\section*{Compliance with Ethical Standards}

This study was funded by the National Natural Science Foundation of China Grant 61370142 and Grant 61272368, by the Fundamental Research Funds for the Central Universities Grant 3132016352, by the Fundamental Research of Ministry of Transport of P.R. China Grant 2015329225300. Huibing Wang, Haohao Li and Xianping Fu declare that they have no conflict of interest. This article does not contain any studies with human participants or animals performed by any of the authors.


\bibliographystyle{unsrt}
\bibliography{Reference}   


%
%

\end{document}